\documentclass[lettersize,journal]{IEEEtran}
\usepackage{amsmath,amsfonts}
\usepackage{algorithmic}
\usepackage{algorithm}
\usepackage{array}
\usepackage[caption=false,font=normalsize,labelfont=sf,textfont=sf]{subfig}
\usepackage{textcomp}
\usepackage{stfloats}
\usepackage{url}
\usepackage{verbatim}
\usepackage{graphicx}
\usepackage{cite}
\usepackage{threeparttable}
\usepackage{xcolor}
\usepackage{multirow}
\usepackage{booktabs}
\usepackage{ragged2e}

\makeatletter
\let\NAT@parse\undefined
\makeatother
\usepackage[hidelinks]{hyperref}
\usepackage{orcidlink}

\begin{document}

\title{A Human-Sensitive Controller: Adapting to Human Musculoskeletal Disorder-Related Constraints via Reinforcement Learning}

\author{
    Vitor Martins\orcidlink{0009-0004-1385-0984},
    Sara M. Cerqueira\orcidlink{0000-0002-8097-5507},
    Mercedes Balcells\orcidlink{0000-0002-2532-0516},
    Elazer R Edelman\orcidlink{0000-0002-7832-7156},
    Cristina P. Santos\orcidlink{0000-0003-0023-7203}
    
    \thanks{This work was supported in part by the Fundação para a Ciência e Tecnologia (FCT) under the national support to R\&D units grant, through the reference project UID/04436: Centro de Microssistemas Eletromecânicos da Universidade do Minho (CMEMS-UMinho), and by Project INTEGRATOR under grant 2022.15668.MIT. Sara Cerqueira was supported by the doctoral Grant SFRH/BD/151382/2021, financed by FCT, under MIT Portugal Program.}
    \thanks{Vitor Martins and Sara M. Cerqueira are with Center for MicroElectroMechanical Systems (CMEMS), University of Minho, Guimarães, Portugal  {\tt\small \{vhugobmartins@gmail.com, saracerqueira1996@gmail.com\}} .}
    \thanks{Mercedes Balcells is with IMES, Massachusetts Institute of Technology, Cambridge, MA, USA and with GEVAB, IQS School of Engineering, Barcelona, Spain {\tt\small \{merche@mit.edu\}}.}
    \thanks{Elazer R Edelman is with IMES, Massachusetts Institute of Technology, Cambridge, MA, USA  and with Brigham and Women’s Hospital, Harvard Medical School Boston, MA, USA {\tt\small \{ere@mit.edu\}}.}
    \thanks{Cristina P. Santos is with Center for MicroElectroMechanical Systems (CMEMS), University of Minho, Guimarães, Portugal and with LABBELS-Associate Laboratory, University of Minho, Guimarães, Portugal {\tt\small \{cristina@dei.uminho.pt\}}.}
}

\maketitle

\begin{abstract}
Work-Related Musculoskeletal Disorders continue to be a major challenge in industrial environments, leading to reduced workforce participation, increased healthcare costs, and long-term disability. This study introduces a human-sensitive robotic system aimed at reintegrating individuals with a history of musculoskeletal disorders into standard job roles, while simultaneously optimizing ergonomic conditions for the broader workforce. This research leverages reinforcement learning (RL) to develop a human-aware control strategy for collaborative robots, focusing on optimizing ergonomic conditions and preventing pain during task execution. Two RL approaches, Q-Learning and Deep Q-Network (DQN), were implemented and tested to personalize control strategies based on individual user characteristics. Although experimental results revealed a simulation-to-real gap, a fine-tuning phase successfully adapted the policies to real-world conditions. DQN outperformed Q-Learning by completing tasks faster while maintaining zero pain risk and safe ergonomic levels,
achieving on average 38\% shorter task completion times across all tested anthropometries. The structured testing protocol confirmed the system’s adaptability to diverse human anthropometries, underscoring the potential of RL-driven cobots to enable safer, more inclusive workplaces.
\end{abstract}


\renewcommand{\thefootnote}{}
\footnotetext{This work has been submitted to the IEEE for possible publication. Copyright may be transferred without notice, after which this version may no longer be accessible.}

\section{Introduction}

Work-related musculoskeletal disorders (WRMSDs) represent a major global health and economic burden. They account for 53\% of all occupational diseases, with estimated annual costs reaching €240 billion in Europe and \$213 billion in the United States \cite{Bevan2015, YELIN2016}. In Europe, 60\% of workers consider pain-related musculoskeletal disorders (MSDs) to be their most serious health issue.

One in three individuals affected by musculoskeletal disorders (MSDs) do not believe they can work until the age of 60. MSDs are responsible for 30\% of all years lived with disability \cite{EU_OSHA_2019} and lead to major productivity losses, increased absenteeism, and early retirement \cite{Bellosta2022, Bernfort2021}, placing a heavy burden on both public health systems and national economies. Australia alone reported annual losses of \$32 billion \cite{EU_OSHA_2019}. The human impact is also significant, with many individuals experiencing frustration and a heightened risk of depression \cite{Shi2016, EU_OSHA_2019}. These challenges are intensified by Europe’s aging workforce, where 67.5\% of workers are nearing retirement \cite{Schramm2016}. This  further increases the incidence of work-related MSDs and raising concerns about future labor shortages \cite{Kugler2022}.

Addressing these challenges requires both preventive strategies and inclusive workplace design. Preventing WRMSDs involves reducing risk factors such as awkward postures. However, when injuries occur, workstations must be adapted to accommodate individual physical limitations. Human-robot collaboration (HRC) offers a promising approach to develop personalized work environments that support healthy and impaired workers. To the best of the authors’ knowledge, the design of controllers that minimize pain-inducing movements while ensuring ergonomic assistance has only been explored in our previous work \cite{Martins2025ourpaper}.

Developing personalized controllers in HRC is a difficult task, since human motion is complex and hard to model. However, several recent studies have demonstrated the potential of reinforcement learning (RL) to enhance human-robot collaboration with a focus on ergonomics and safety, which informed the direction of this work.  \cite{Lagomarsino2023} applied RL to achieve low-effort and predictable handovers;  \cite{Human-robot-gym2024} introduced a simulation platform for benchmarking RL agents and safety mechanisms; and \cite{Xie2024} combined 3D skeleton reconstruction with online RL to dynamically reduce ergonomic risk from awkward postures.

In our previous work \cite{Martins2025ourpaper}, we introduced a RL-based control strategy for human-robot collaboration, demonstrated through a collaborative object transport task. This controller was specifically developed to support the reintegration of individuals recovering from musculoskeletal disorders by minimizing pain-inducing movements and adapting to individual physical limitations, while also benefiting the general workforce by continuously optimizing ergonomic conditions. This approach relied on a Q-learning algorithm implemented within a discretized action space, where the robot's movements were constrained to fixed step lengths of 10 cm and 6.5 cm, corresponding to the discretization granularity. 
Although the approach yielded promising results in ergonomic performance and indicated no pain risk, it was validated with only one participant, limiting the generalizability of the findings.
Expanding this work, the current work aims to benchmark the initial Q-learning strategy against a Deep Q-Network (DQN)-based controller, which introduces variable step lengths and directions to potentially improve task efficiency. Additionally, we seek to evaluate the robustness and adaptability of the approach by extending testing to subjects with diverse physical profiles.

The contributions of this work are threefold: (1) development of a DQN-based controller with variable step sizes and integrated action shaping, enabling flexible, efficient behavior while learning user-specific ergonomic needs and physical constraints; (2) real-world testing and benchmarking against a prior Q-learning approach, demonstrating superior performance in task efficiency and user comfort; and (3) validation with participants of diverse anthropometries, confirming the controller’s adaptability and generalizability.

 \section{Solution Conceptualization}\label{sec:Solution Conceptualization}
 \subsection{Solution requirements}

This work builds upon the initial conceptualization of our prior work \cite{Martins2025ourpaper}, where we defined a human-sensitive control strategy to support collaborative object transport. The controller was designed to meet three primary requirements: (i) pain-free operation, by avoiding movements that could induce discomfort; (ii) ergonomic assistance, by minimizing awkward postures; and (iii) task efficiency, by completing the transport task within minimal time.

\subsection{Elbow Contracture and Use Case Scenario}

A key challenge in occupational rehabilitation is reintegrating individuals with musculoskeletal disorders into standard job roles without requiring task redesign or workstation modifications. This work addresses that challenge by developing a robotic controller that learns to respect the worker's physical constraints through pain and ergonomic risk assessment, enabling the system to proactively adapt its behavior to individual capabilities rather than demanding adaptation from the worker.

\begin{figure}
 \centering
 \includegraphics[scale=0.30]{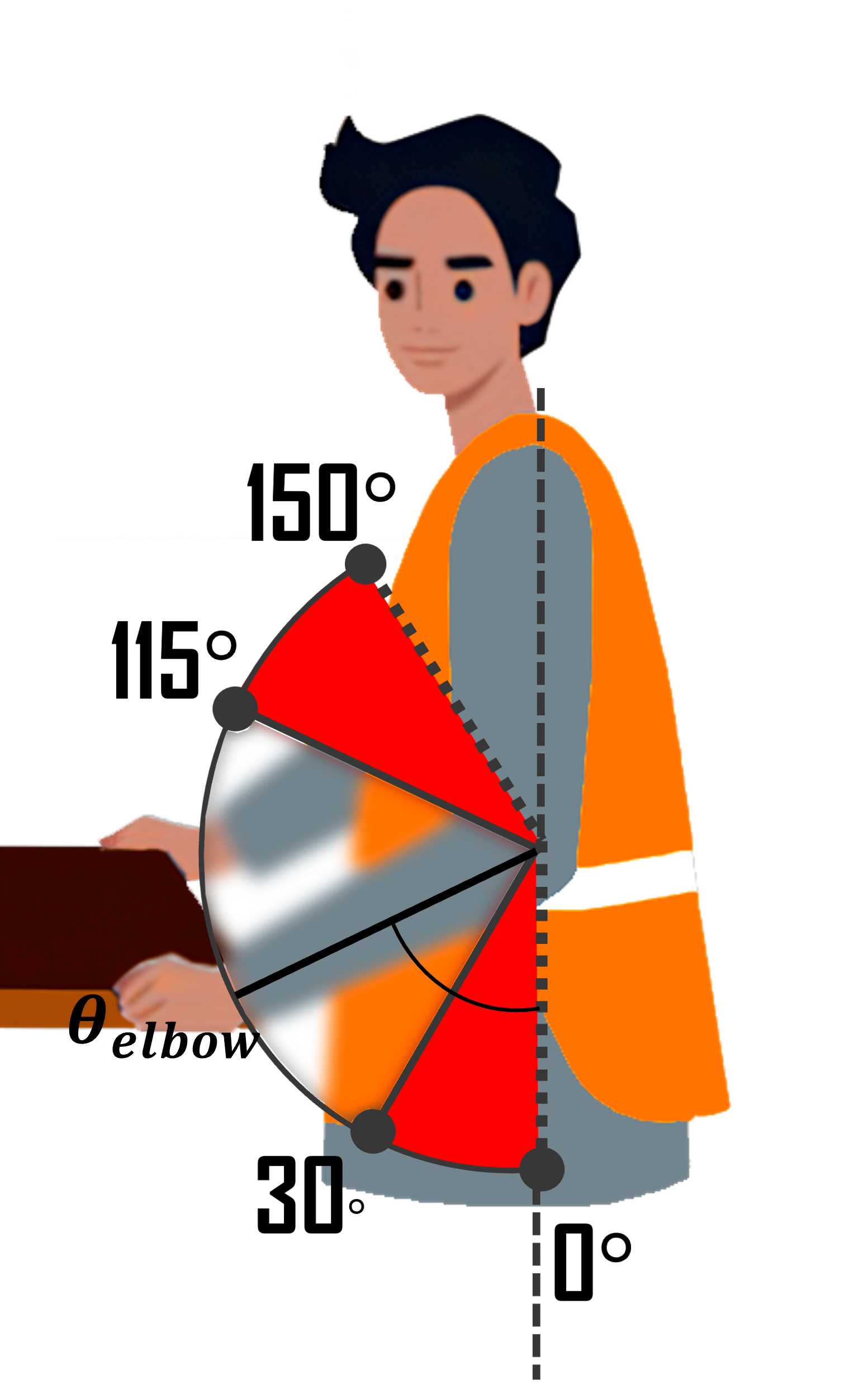}
 \caption{Elbow contracture ROM. Red represents the painful elbow's ROM. }
\label{fig: Elbow contracture ROM}
\end{figure}

Although the goal is to develop a broadly applicable solution for a variety of MSDs through the use of digital pain biomarkers, this exploratory study focuses on a representative use case: elbow contracture. This condition limits elbow range of motion \cite{Masci2020}, typically restricting extension beyond 30° and flexion below 120°, and can cause pain within the affected arc as illustrated in Fig. \ref{fig: Elbow contracture ROM}. 

As a proof-of-concept, this study focuses on a co-manipulation task,  where a human and robot jointly transported an object vertically. The task endpoints were tailored to the participant’s anthropometric limits.

\section{Reinforcement Learning Design} \label{sec:Materials and Methods}

\subsection{Train Methodology}
Training RL agents entirely in real-world scenarios is often unfeasible due to the extensive time requirements and the difficulty of engaging participants in prolonged sessions. In contrast, the stochastic and unpredictable nature of human behavior limits the fidelity of simulation-only approaches for modeling human-centered interactions

To overcome the limitations of purely simulated or real-world training, a two-stage pipeline was implemented that integrates both domains. Initially, the agent is pre-trained in a simulated environment to acquire fundamental behaviors and minimize the demands of real-world data collection. A kinematic human model, individually scaled based on participant anthropometry, is employed to perform inverse kinematics (IK) and estimate ergonomic and pain-related risks from simulated postural responses. After this pre-training phase, the learned policy is transferred to the real-world setting for fine-tuning and personalization. During this phase, a sensory system monitors the human collaborator’s posture to support real-time policy adaptation. To ensure applicability across individuals, the human model is scaled to the anthropometric characteristics of each participant.

\subsection{Reinforcement Learning Formulation}

In our previous work\cite{Martins2025ourpaper}, Q-Learning was selected for its conceptual simplicity, model-free nature, and fast convergence in well-defined Markov Decision Processes (MDPs). These features are especially useful in human-in-the-loop settings, where fast training is essential due to limited participant availability and tolerance for long sessions. While effective for solving RL problems modeled as simple MDPs, traditional Q-Learning is limited by its reliance on discrete state-action representations, which hinder scalability in more complex or continuous environments. To address this, we implemented Deep Q-Networks (DQN), which extend Q-learning by using neural networks to approximate the Q-function. 
Although DQN incurs higher computational complexity from its neural network updates and replay buffer during initial simulation pre-training compared to tabular Q-Learning's simple table updates and its ability to handle continuous state representations unlocks new possibilities for performance enhancement, such as finer-grained object positioning. While more advanced RL methods exist, DQN enables effective generalization in high-dimensional state spaces and facilitates faster training, making it well-suited for real-world applications where computational resources and time are often constrained.
An $\epsilon$-greedy strategy was used for action selection, to balance exploration and exploitation. With DQN, the controller remains lightweight and sample-efficient, making it well-suited for real-time training with human users while scaling to more realistic interaction settings.

The RL paradigm was modeled as a fully observable MDP. The key components of the RL environment were defined as follows (see Fig. \ref{fig:geral}):

\begin{figure}[thpb]
    \centering
    \includegraphics[scale=0.33]{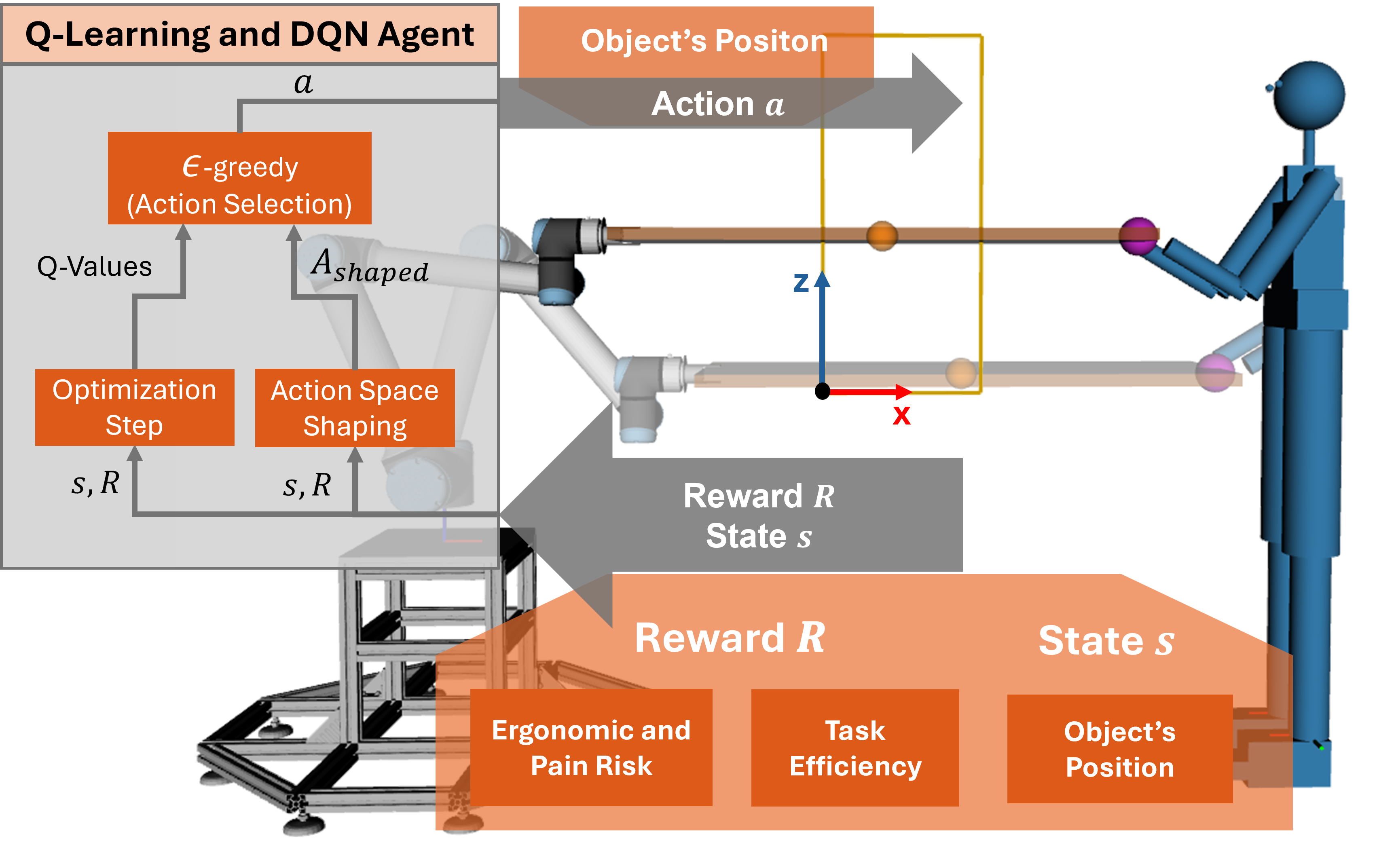}
    \caption{RL framework overview for learning human specificities, in ergonomics and physical constraints.}
    \label{fig:geral}
\end{figure}

The \textbf{state space} $S$ was defined as the Cartesian area where the object can be displaced. In our prior work, the Q-Learning state space was discretized using a resolution factor $\Delta_d$, where each state $s \in S$ represents the discretized cell coordinate containing the object's position. In contrast, the DQN state directly utilized the object’s Cartesian position, defined as the midpoint between the end-effectors (EE) of the Cobot and the Human. This position was encoded as a two-dimensional vector $(x_{\mathrm{obj}}, z_{\mathrm{obj}})$, where $x$ and $z$ denote the horizontal and vertical coordinates, respectively, according to the reference frame in Figure \ref{fig:geral}. 

The \textbf{action space} $A$ defined how the agent interacts with the environment by modifying the object's position. Both Q-Learning and DQN employ discrete action representations. Previously, the Q-Learning action space was defined by movements between grid cells. To allow applicability to a continuous state space domain, the DQN action was represented as a pair $(\alpha, \textit{d})$, where $\alpha$ denotes the movement direction and $d$ represents the distance traveled. The possible values for $\alpha$ are \{0, 30, 60, 90, 120, 150, 180\} degrees, and for $d$ are \{0.02, 0.03, 0.05, 0.2, 0.4\} meters. These values offer a balance between directional flexibility and step size variability. The chosen angles avoid backward movements, while multiple distance options allow the agent to plan more efficient transport trajectories by potentially reducing the number of required steps.

The \textbf{reward function} $\mathrm{R}$ was designed to guide the agent's behavior by balancing three key objectives: minimizing pain risk, reducing ergonomic risk, and promoting task efficiency. Actions that lead to high ergonomic or pain risks are penalized, while those that improve ergonomic conditions, mitigate pain risks, and contribute to task progress are rewarded.

An \textbf{episode} consists of a sequence of state-action transitions, beginning with the object in its initial position and ending when it reaches a predefined target area (the top of the cartesian area).

\section{Experimental Setup and Implementation Details} \label{sec: Experimental Setup}

The setup features a 6 degrees-of-freedom (DOF) industrial collaborative robot (UR10e, Universal Robot). For the co-manipulation task, a wooden board was attached to the gripper's coupling system. 
In the simulation environment, the URSim simulator (Universal Robots) was used together with a human kinematic model integrated within the same ROS control loop to emulate robot behavior under the selected RL actions and to simulate the corresponding human posture configuration. Since only kinematics were relevant for the training process, RViz was used as the visualization and simulation platform, providing a sufficient environment without the need for more complete physics-based simulators. 

In the real-world environment, joint-angle data is acquired using the MVN Awinda motion capture system (Movella Technologies B.V., Netherlands), operating at 60 Hz. Data is transmitted to the control framework via UDP and processed by a kinematic model scaled to the subject’s anthropometry. The model computes the Cartesian position of the human hands or EE using direct kinematics.
Figure \ref{fig: environments} shows the experimental setup in both simulated and real-world environments.

\subsection{Simulation Environment and Human Posture Modeling}   \label{subsec: Simulation Environment and Human Posture Modeling}
The challenge in the simulation environment is to accurately model the Human body configuration, after the robot starts the object displacement. To address this, a kinematic model scaled to the Human's anthropometries was defined. 
Given the vertical transport task, primarily involving shoulder and elbow motion, the kinematic chain was modeled as a 2-DOF system. Assuming symmetric limb movement in the sagittal plane, the model captures the key dynamics of the task, specifically, the flexion-extension of the shoulder and elbow joints.

Joint limits were defined according to normative shoulder and elbow flexion-extension ranges of motion (ROM) \cite{Elisa, physiopediaShoulderROM}: shoulder angles $\theta_{\mathrm{shoulder}} \in [-60^\circ, 180^\circ]$ and elbow angles $\theta_{\mathrm{elbow}} \in [0^\circ, 150^\circ]$.
Inverse kinematics was used to compute the shoulder and elbow joint angles. When the RL agent proposes a new Cartesian object position $(x_{\mathrm{des}}, z_{\mathrm{des}})$, the corresponding human EE position is estimated as $(x_{\mathrm{des}} + \frac{L_{\mathrm{object}}}{2}, z_{\mathrm{des}})$, accounting for half the object’s length along the x-axis.
Given the 2-DOF kinematic model and the computed human EE position, inverse kinematics provides two possible solutions. The solution that is physically feasible and aligns with the human's natural configuration, $(\theta_{\mathrm{shoulder}}, \theta_{\mathrm{elbow}})$, is selected. 

\subsection{Participants}   \label{subsec: Participants}

This exploratory study was conducted with four healthy participants (3 male and 1 female). They were recruited to ensure diversity in height to enable evaluation across different anthropometric profiles. The participants ranged in age from 22 to 48 years (29.25 $\pm$ 12.53) and had an average height of 1.713 $\pm$ 0.107 meters. The 2-DOF model was scaled for each individual using their height, shoulder span, upper-arm length, and forearm length.

\begin{figure}[thpb]
     \centering
     \includegraphics[scale=0.5]{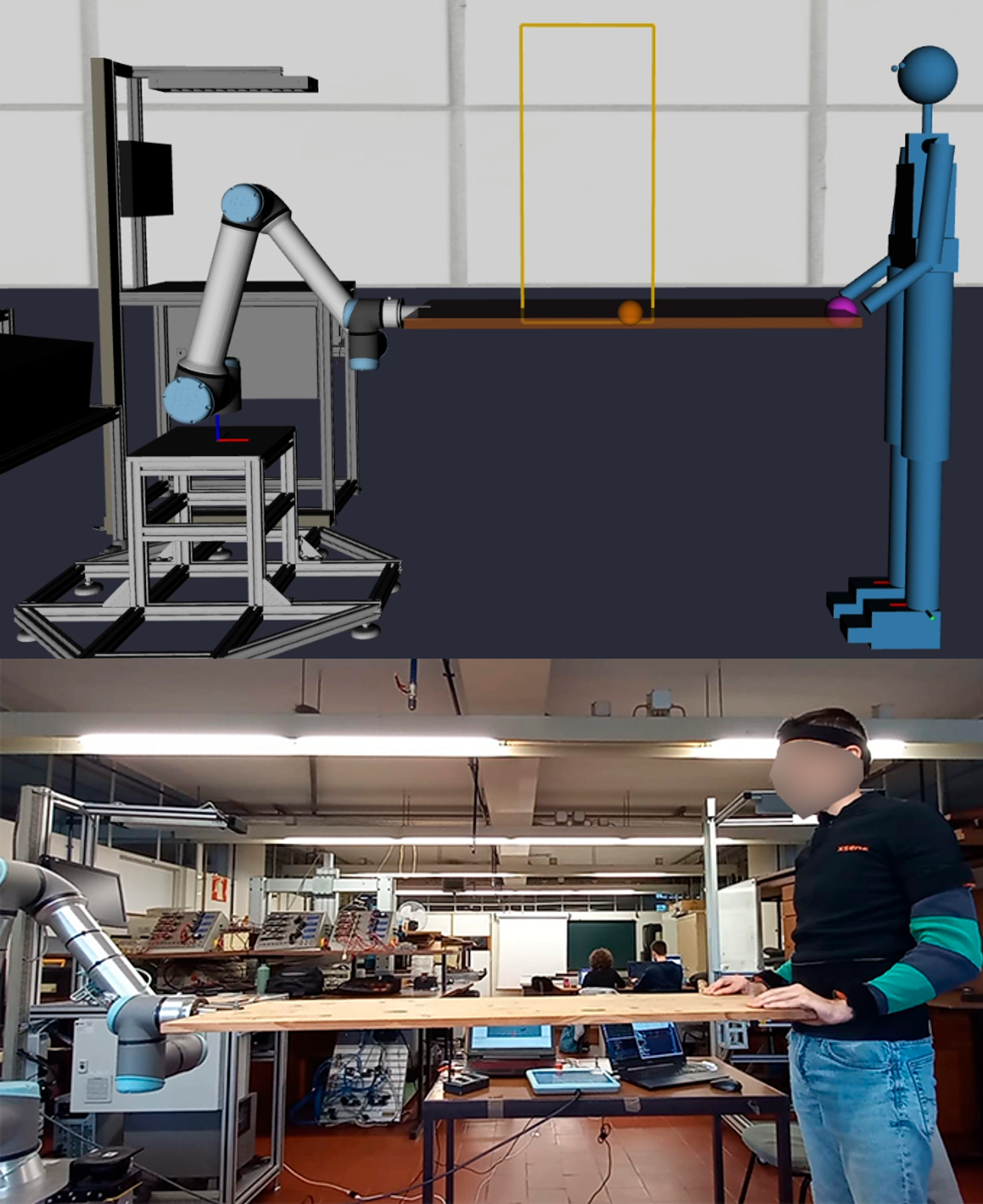}
     \caption{Simulation (top) and real-world (bottom) environments. In the real-world environment, the participant is equipped with Movella’s upper-body motion capture system.}
     \label{fig: environments}
     \vspace{-5pt} 
\end{figure}

All participants were instrumented using an upper-body motion capture system. A total of nine inertial wireless trackers were placed according to the manufacturer’s guidelines, specifically on the head, shoulders, chest, upper arms, forearms, and waist, and secured using Velcro straps (Fig. \ref{fig: environments}). The system was calibrated in the “upper body, no hands” configuration. Calibration was repeated as needed until visual inspection confirmed satisfactory tracking accuracy.
This study followed the ethical guidelines of the CEICVS (147/2021), the Declaration of Helsinki, and the Oviedo Convention.

\subsection{State and Action Space Details}   \label{subsec: State Space Detail}
The state space is defined in a 90 cm (height) × 40 cm (width) Cartesian area, located 1 meter from the robot base. All participants stood 2.72 meters from the base with fixed foot positions.
Initial and target transport positions were customized per participant to ensure reachability, based on anthropometric assessments.

Table \ref{tab: initial_state_end_condition} summarizes the initial and target conditions for each participant and algorithm. In Q-Learning, states are represented by grid positions, whereas in DQN, they are defined in Cartesian coordinates  ($x$, $z$ in meters). 
To ensure reachability for all participants, both initial and target conditions are personalized based on individual anthropometry. In Q-Learning, success is defined as reaching a specific target height ($z_{\mathrm{obj}}$). In DQN, success is achieved when $z_{\mathrm{obj}} \geq 90 - \Delta z$, where $\Delta z$ is a participant-specific height margin.

\begin{table}[htpb]
    \centering
    \caption{Initial States and End Episode Conditions}
    \label{tab: initial_state_end_condition}
    \renewcommand{\arraystretch}{1.2} 
    \begin{tabular}{c c c c c } 
        \toprule
                       & \multicolumn{2}{c}{\textbf{Initial State}} & \multicolumn{2}{c}{\textbf{End Condition}} \\ \cmidrule(lr){2-3} \cmidrule(lr){4-5}
           Participant's Height & QL$^{1}$    & DQN             & QL ($z_{\mathrm{obj}}$)              & DQN ($\Delta z$)               \\ \midrule
        1.62           & (4, 1)& (1.3, 0.434)& 9   & 0.20 \\    
        1.69           & (4, 1)& (1.35, 0.454)& 9   & 0.15  \\ 
        1.79           & (4, 1)& (1.3, 0.434)& 10  & 0.03  \\
        1.83           & (5, 2)& (1.35, 0.474)& 10  & 0.03 \\ \bottomrule
    \end{tabular}
    \begin{tablenotes}
        \item $^{1}$Q-Learning
    \end{tablenotes}
\end{table}

Each RL action is executed by the UR10e robot. The EE target position is computed from the desired Cartesian position of the object, $(x_{\mathrm{des}}, z_{\mathrm{des}})$, and horizontally offset by 0.65 meters (half the object's length), resulting in a final EE target of $(x_{\mathrm{des}} - 0.65, z_{\mathrm{des}})$.

\subsection{Action Space Shaping}

In prior work\cite{Martins2025ourpaper}, action space shaping was used to improve learning efficiency by restricting the agent’s available actions at each step. Specifically, the Q-Learning agent was prevented from selecting actions that moved the object backward. In the proposed DQN formulation, this constraint is inherently satisfied through the directional limits of the action space ($\alpha \in [0^\circ, 180^\circ]$), eliminating the need for additional checks. 
However, DQN requires validation to ensure that actions stay within the state-space boundaries. This is done by calculating the new position $(x_{\mathrm{des}}, z_{\mathrm{des}})$ using the current object position $(x_{\mathrm{obj}}, z_{\mathrm{obj}})$ and the action vector $(d \cdot \cos{\alpha}, d \cdot \sin{\alpha})$. The computed position must satisfy $x_{\mathrm{des}} \in [0, x_{\mathrm{max}}]$ and $z_{\mathrm{des}} \in [0, z_{\mathrm{max}}]$. Actions that violate these bounds are discarded from the available action set $A$.

In addition, all actions must comply with the kinematic model. This includes verifying that the simulated human EE position, derived from the desired object's position, remains within the kinematic workspace. An action is deemed valid only if it has a feasible IK solution and adheres to joint limit constraints. This validation process is applied in both simulated and real environments, resulting in a shaped action space, $A_{\mathrm{shaped}}$, which guarantees the selection of only safe and efficient actions. This approach is particularly beneficial in real-world applications, as it prevents scenarios where selected actions cannot be physically executed by the human. 

Further, an episode may terminate before reaching target condition if the shaped action space $A_{\mathrm{shaped}}$ becomes empty (e.g., no valid actions satisfy the IK constraints), or if a pain risk is detected in the real environment.

\subsection{Pain and Ergonomic Assessment}   \label{subsec: Assessment Methods}

Ergonomic risk is assessed using the Rapid Upper Limb Assessment (RULA) method \cite{MCATAMNEY199391}, a widely used tool in ergonomic analysis. As the task primarily involves shoulder and elbow movement (see Section \ref{subsec: Simulation Environment and Human Posture Modeling}), the RULA ergonomic score typically falls within the integer set {1, 2, 3}. 
Scores of 1 and 2 indicate acceptable postural conditions, with risk increasing progressively with the score. A score of 3 signals an elevated postural load, warranting further investigation and corrective postural adjustments.

Pain assessment is designed to eventually integrate a digital pain biomarker, a model capable of continuously predicting, in real time, whether a person is experiencing pain, currently under development by the research team. In this proof of concept, the biomarker is abstracted through a state machine module, which serves as a placeholder within the control framework.
In this study, pain is estimated based on elbow range-of-motion (ROM) constraints, which are key indicators of discomfort (Figure \ref{fig: Elbow contracture ROM}). A state machine monitors the elbow joint angle and returns a binary pain state: a pain state of "1" is assigned when the angle falls within [0°, 30°] or [115°, 150°], and "0" otherwise. The module is encapsulated to allow easy replacement with a digital biomarker in future work.
Real-time joint angle measurements provided by the motion capture system serve as inputs to the RULA method and the state machine module, enabling the computation of ergonomic and pain risk levels, respectively.

In the real environment, $\mathrm{avgErg}$ and $\mathrm{avgPain}$ are computed by averaging the ergonomic and pain risk levels received at the sensory system’s update rate during object movement. In simulation, the UR10e’s linear motion is discretized along the path from the initial to the desired object position. At each discretized point, inverse kinematics is applied to compute the corresponding risk levels. The final $\mathrm{avgErg}$ and $\mathrm{avgPain}$ are obtained by averaging the values across all points along the trajectory, starting from the initial state.

\subsection{Reward Function}
The reward function in DQN extends the Q-Learning reward structure from our previous work. It incorporates an ergonomic reward term ($\mathrm{ergRew}$), which penalizes high ergonomic risk levels and is defined as:

\begin{equation}
    \mathrm{ergRew} = -50 \, \mathrm{avgErg} + 50
\label{eq:ergRew}
\end{equation}
where $\mathrm{avgErg}$ $\in [1, 3]$ represents the average ergonomic risk level after action execution.

Given the task involves lifting an object vertically, according to ergonomic guidelines, an increase in ergonomic risk is unavoidable.
As a result, the agent is tempted to perform sideward movements to dilute the long-term average ergonomic risk level. However, this introduces unnecessary movements, reducing time efficiency. To mitigate this, it was included into the reward function ($\mathrm{xMovRew}$, equation \ref{eq:xMovRew}) a component that discourages sideward movements $(x \pm 1, z)$, i.e., consecutive movements ($\mathrm{countX}$) along the horizontal axis are penalized linearly.

\begin{equation}
\mathrm{xMovRew} =
\begin{cases} 
-20 \, \mathrm{countX}, & \mathrm{countX} \leq 5 \\
-100, & \mathrm{countX} > 5
\end{cases}
\label{eq:xMovRew}
\end{equation}

DQN's reward function  adds a new component, $\mathrm{zStepRew}$, when the pain risk is zero. This component rewards the agent based on the magnitude of the vertical movement ($z$-axis), encouraging time-efficient behavior and discouraging unnecessary adjustments that offer minimal ergonomic benefit. 
The reward for vertical movement is given by:
\begin{equation}
\mathrm{zStepRew} = \frac{\mathrm{d} \times \sin{\alpha} \times 100 }{0.4} 
\label{eq:zStepRew}
\end{equation}

This component is also normalized between 0 and 100. The complete reward function is defined in Equation~\ref{eq:RewardDQN}, where the weights of each term were selected through trial and error.

\begin{equation}
\mathrm{R} =
\begin{cases} 
    -100, & \mathrm{avgPain} = 1 \\
    0.15 \, \mathrm{ergRew} + 0.3 \, \mathrm{zStepRew}  \\
    + 0.05 \, \mathrm{xMovRew}, & \mathrm{avgPain} = 0
    \end{cases}
\label{eq:RewardDQN}
\end{equation}

This reward structure ensures that the agent is penalized for high pain and ergonomic risks, while incentivizing movements that reduce these risks and advance the transport task. Components such as  $\mathrm{ergRew}$, $\mathrm{xMovRew}$, and the total reward $\mathrm{R}$ are normalized between -100 and 0 to maintain consistent scaling. These elements follow the generic RL framework depicted in Figure \ref{fig:geral}. 

In DQN, the Q-Table is replaced by a NN, implemented as a fully connected network (FCN). The input layer receives the object’s position, $x_{\mathrm{obj}}$ and $z_{\mathrm{obj}}$, normalized to the range $[-1, 1]$ based on the state space limits. ReLU activation functions are used in all layers. The network includes a single hidden layer (size specified in Section~\ref{subsec: res_hpo}) with ReLU activation. The output layer consists of 35 neurons (the size of the discrete action space) and outputs Q-values.

\subsection{Training Termination Criteria} \label{subsec: Train Methodology}
For the training settings, the termination criteria are different between the simulation and the real-world environment. In the simulation, training concludes when the moving average of the reward per episode remains constant over 100 episodes. In the real-world environment, training ends when the average of the ergonomic risk level remains below 2.5  with zero pain risk for 10 consecutive episodes. The ergonomic risk level threshold reflects RULA guidelines, which recommend further investigation and possible changes to the workspace for scores above 3.

\subsection{Testing Protocol}
To ensure optimal performance, the following procedure was followed:

\begin{enumerate} \item \textbf{Hyperparameter Optimization (HPO):} Models were pre-trained and evaluated in simulation based on reward, pain and ergonomic risk, and time efficiency. The best-performing model was selected. \item \textbf{Real-World Deployment:} The selected model was tested over 10 episodes in the real environment, with key metrics recorded. \item \textbf{Fine-Tuning:} The model was adapted to real-world dynamics through fine-tuning. \item \textbf{Final Evaluation:} The fine-tuned model was re-tested in the real environment to assess performance gains. \end{enumerate}

This protocol was repeated for each participant and both algorithms, except for the HPO step, which is performed only for DQN. 
The system's verification is primarily based on the defined performance metrics derived from human joint angles. To further support objective findings and provide qualitative confirmation of the system’s real-world behavior, each participant completed a usability questionnaire at the end of the experiments. The questionnaire was designed to capture subjective perceptions of comfort, safety, and potential industrial applicability.

The participants were asked to respond to the following four questions 
(Q1 directly compares Q-Learning and DQN comfort; Q2-Q4 assess the fine-tuning phase's impact on comfort/safety for both algorithms and the controller's overall industrial usability):
\begin{itemize}
    \item \textbf{Q1}: Which experiment was more comfortable?
    \item \textbf{Q2}: I felt that, after fine-tuning the model, the transport became more comfortable.
    \item \textbf{Q3}: I felt that the fine-tuning phase helped keep the elbow angle within safe limits, avoiding excessive flexion or extension.
    \item \textbf{Q4}: I believe this controller would be useful in industrial settings, enabling adaptation to individual physical needs, including WRMSD workers.
\end{itemize}

\subsection{Hyperparameter Optimization}   \label{subsec: Hyperparameter Optimization}
Hyperparameter optimization (HPO) aims to identify the best-performing model but is time-consuming, as it would need to be repeated separately for each participant. To reduce overhead, HPO for DQN was performed only once, using the same human kinematic model employed in the Q-Learning testing protocol. The resulting hyperparameter set was then used for pre-training and fine-tuning DQN models for the remaining participants.

For DQN HPO,  a grid search was conducted over the parameter space defined in Table~\ref{tab:hyperparameters}, totaling 2187 experiments. To improve efficiency, an early stopping criterion was applied: training was terminated if the duration exceeded twice the duration of the epsilon decay. This hyperparameter controls the exploration-exploitation balance by regulating the linear decay of epsilon to zero, defining the number of episodes required for complete epsilon decay.

\begin{table}[h]
    \centering
    \caption{DQN Hyperparameters Search Space}
    \label{tab:hyperparameters}
    \begin{tabular}{ c c }
        \hline
        \textbf{Hyperparameter}         & \textbf{Search Space}     \\ \hline
        Learning Rate                   & {1e-5, 1e-4, 1e-3}\\ 
        Discount Factor                 & {0.5, 0.9, 0.999}\\ 
        Epsilon Decay Duration          &  {1500, 2000, 2500}\\ 
        Soft Update Rate                & {1e-4, 1e-3, 1e-2}\\ 
        Buffer Size                     & {5000, 10000, 20000}\\ 
        Batch Size                      & {64, 128, 256}\\ 
        Neurons No.                     & {128, 256, 512}\\ \hline
    \end{tabular}
\end{table}

\section{Results and Discussion}   \label{sec: Results and Discussion}
    \subsection{DQN Hyperparameter Optimization} \label{subsec: res_hpo}
Two machines were used to perform the HPO, with the following specifications: Machine A) GPU: Nvidia GeForce GTX 1650 (dedicated memory: 4 GB); CPU: AMD Ryzen 7 4800H with Radeon Graphics 2.90 GHz (8 cores, 16 threads); RAM: 16 GB; and Machine B) GPU: NVIDIA GeForce RTX 3080 Ti (Dedicated memory: 12 GB); CPU: Intel Core i9-10940X CPU 3.30 GHz (14 cores, 28 threads); RAM: 62 GB. Machine A trained 1207 experiments in 387 hours, while machine B trained 980 experiments and took 336 hours to complete. The best-performing model was obtained with a Learning Rate of 1, a Soft Target Update Rate of $1 \times 10^{-3}$, Discount Factor of 0.999, 512 neurons on the FCN's hidden layer, a Batch Size of 64 transitions, a Buffer Size of 5000 and a Epsilon Decay Duration of 1500 episodes.

The results of the best-performing model from the HPO for each algorithm are summarized in Table \ref{tab:bestModelsHPO}. 
\begin{table}[h!]
    \centering
    \begin{threeparttable}
        \caption{Best-performing models results}
        \label{tab:bestModelsHPO}
        \begin{tabular}{l c c c c c } \hline 
            \textbf{Algorithm} &\textbf{Reward}&  \textbf{Pain$^{1}$}&  \textbf{Erg$^{2}$}& \textbf{Steps} &\textbf{Ep. Conv.$^{3}$}\\ \hline 
            QL&-70&  0&  2.00& 9 & 195\\ 
            DQN& 67& 0& 2.12& 5&1609\\ \hline
        \end{tabular}
        \begin{tablenotes}
            \item $^{1, 2}$ Average pain and ergonomic risk levels, respectively; $^{3}$ Episode in which the reward started to converge.
        \end{tablenotes}
    \end{threeparttable}
\end{table}
 
The reward values obtained by the DQN model are not directly comparable to those of Q-Learning, as they are based on different reward functions (Eq. \ref{eq:RewardDQN}). However, when comparing other performance metrics, both models achieved zero pain risk. Although DQN exhibited a slightly higher average ergonomic risk level than Q-Learning, the difference was minimal.
A key advantage of the DQN model is its ability to find a solution in just 5 steps, substantially fewer than the Q-Learning model. This efficiency is attributed to the variable step size ($d$) in DQN action space, which allow the task to progress more efficiently. Notwithstanding, this came at the cost of requiring significantly more episodes to reach convergence.

\begin{table*}[]
    \centering
    \begin{threeparttable}
        \caption{Performance Comparison of Q-Learning and DQN Algorithms for Different Human Anthropometries}
        \label{tab:overall_results}
        \begin{tabular}{|c|c|c|c|c|c|c|c|c|}
            \hline
            \textbf{Body Height} & \textbf{Discretization} & \textbf{Train / Exec} $^1$ & \textbf{Reward} & \textbf{Pain Risk}$^2$ & \textbf{Erg. Risk}$^3$ & \textbf{Steps} & \textbf{Distance (m)}$^4$ & \textbf{Time (s)}$^5$ \\
            
            \hline
            \multirow{6}{*}{1.62}   &           & Sim/Sim        & -97.17       & 0  & 2.25           & 9            & 1.07     & ---              \\
                                    &   QL      & Sim/Real       & -145$\pm$49  & 1  & 2.48$\pm$0.21   & 6.6$\pm$3.9  & 0.776$\pm$0.47   & 19.8$\pm$11    \\
                                    &           & Real/Real      & -106$\pm$6   & 0  & 2.42$\pm$0.032  & 8$\pm$0.0    & 0.954$\pm$0.03   & 23.7$\pm$0.0    \\
            \cline{2-9}              
                                    &           & Sim/Sim        & 41          & 0  & 2.14           & 5           & 0.80    & ---              \\
                                    &   DQN     & Sim/Real       & -93$\pm$48  & 1  & 2.19$\pm$0.13  & 7.4$\pm$3.5  & 0.763$\pm$0.02   & 16.3$\pm$1.8     \\
                                    &           & Real/Real      & 25$\pm$6    & 0  & 2.39$\pm$0.04  & 4.5$\pm$1.2  & 0.728$\pm$0.06   & 15.6$\pm$2.0    \\

            \hline \hline
            \multirow{6}{*}{1.69}   &           & Sim/Sim        & -68         & 0  & 2.12           & 8                   & 0.924   & ---              \\
                                    &   QL      & Sim/Real       & -145$\pm$24 & 1  & 2.51$\pm$0.06  & 8$\pm$0.0            & 0.932$\pm$0.01   & 22.1$\pm$0.1     \\
                                    &           & Real/Real      & -80$\pm$12  & 0  & 2.31$\pm$0.08  & 8$\pm$0.0            & 0.916$\pm$0.02   & 21.4$\pm$0.4     \\
            \cline{2-9}             
                                    &           & Sim/Sim        & 45          & 0  & 2.05           & 3                   & 0.83                & ---              \\
                                    &   DQN     & Sim/Real       & -65$\pm$15  & 1  & 2.29$\pm$0.06  & 3.7$\pm$0.5  & 0.852$\pm$0.02   & 15.2$\pm$1.7     \\
                                    &           & Real/Real      & 37$\pm$12   & 0  & 2.33$\pm$0.02  & 3$\pm$0.0            & 0.843$\pm$0.01   & 14.1$\pm$1.4     \\

            \hline \hline
            \multirow{6}{*}{1.79}   &          & Sim / Sim      & -70          & 0  & 2.00           & 9                    & 1.07            & ---             \\
                                    &    QL    & Sim / Real     & -144$\pm$40  & 1  & 2.31$\pm$0.03  & 9$\pm$0.0            & 1.07$\pm$0.0    & 24$\pm$0.0              \\
                                    &          & Real / Real    & -107$\pm$24  & 0  & 2.33$\pm$0.12  & 9$\pm$0.0            & 1.15$\pm$0.0    & 26$\pm$0.0              \\
            \cline{2-9} 
                                    &          & Sim / Sim      & 66.98        & 0  & 2.12             & 5         & 0.92             & ---             \\
                                    &    DQN   & Sim / Real     & -62$\pm$9    & 1  & 2.17$\pm$0.09  & 5.3$\pm$0.5  & 0.91$\pm$0.02   & 16.5$\pm$1.1    \\
                                    &          & Real / Real    & 62$\pm$2     & 0  & 2.22$\pm$0.07  & 4.5$\pm$0.5  & 1.06$\pm$0.01   & 18.4$\pm$1      \\

            \hline \hline
            \multirow{6}{*}{1.83}   &           & Sim/Sim        & -57.390      & 0  & 2.09            & 8$\pm$0.0           & 1.05   & ---             \\
                                    &   QL      & Sim/Real       & -106$\pm$21  & 0  & 2.43$\pm$0.08  & 8$\pm$0.0            & 1.03$\pm$0.0     & 24.3$\pm$0.2    \\
                                    &           & Real/Real      & -87$\pm$5    & 0  & 2.23$\pm$0.04  & 9$\pm$0.0            & 1.02$\pm$0.0     & 26.3$\pm$0.5    \\
            \cline{2-9}
                                    &           & Sim/Sim        & 65.760       & 0  & 2.15            & 3                   & 0.854             & ---             \\
                                    &   DQN     & Sim/Real       & -89$\pm$13   & 1  & 2.45$\pm$0.07  & 3.9$\pm$0.3           & 0.77$\pm$0.04     & 13.5$\pm$1.4    \\
                                    &           & Real/Real      & 92$\pm$4     & 0  & 2.28$\pm$0.08  & 4$\pm$0.0            & 0.75$\pm$0.01     & 12.1$\pm$1.6    \\
            \hline
        \end{tabular}
        \begin{tablenotes}
            \item $^1$Indicates the training and execution environments; $^{2, 3}$Average pain and ergonomic risk levels; $^{4}$Object's traveled distance; $^{5}$Time taken to complete the task in real-world environment.
        \end{tablenotes}
    \end{threeparttable}
\end{table*}

\subsection{Real Environment Tests}

Table~\ref{tab:overall_results} summarizes the performance metrics for both Q-Learning and DQN, with values averaged over 10 real-world trials, as well as the metrics obtained from the execution of the pre-trained models in the simulated environment.

The pre-training phase successfully produced policies that enabled efficient transport, maintained ergonomic risk below 2.5, and avoided pain risk. However, transitioning from simulation to real-world execution revealed a significant simulation-to-real gap, with reward values dropping across all experiments and participants. This decline stemmed from the emergence of pain risk and increased ergonomic risk. 
The gap likely results from a combination of factors, including simplifications in the human kinematic model and limitations of the sensory system, such as potential calibration mismatches or magnetic interference.

The DQN pre-training phase took, on average, 15 minutes longer than Q-Learning (19.2 min vs. 4.2 min). However, during fine-tuning, DQN converged 6 minutes faster (5.4 min vs. 11.4 min). In terms of transport efficiency, measured by steps, object distance, and time per episode, fine-tuned models performed similarly to their simulated counterparts. Notably, DQN achieved the most efficient transport, completing it in just 3 steps, 5 fewer than the best fine-tuned Q-Learning model (8 steps), resulting in a 7.3 second reduction in transport time (14.1 s vs. 21.4 s).

Overall, fine-tuning the policy in the real-world significantly improved performance, bringing reward values close to those in simulation, eliminating pain risk (as the agent learned to avoid joint angle configurations associated with elbow contracture constraints), and keeping ergonomic risk below 2.5, ultimately meeting the solution’s requirements.

\begin{figure}[thpb]
    \centering
    \includegraphics[scale=0.525]{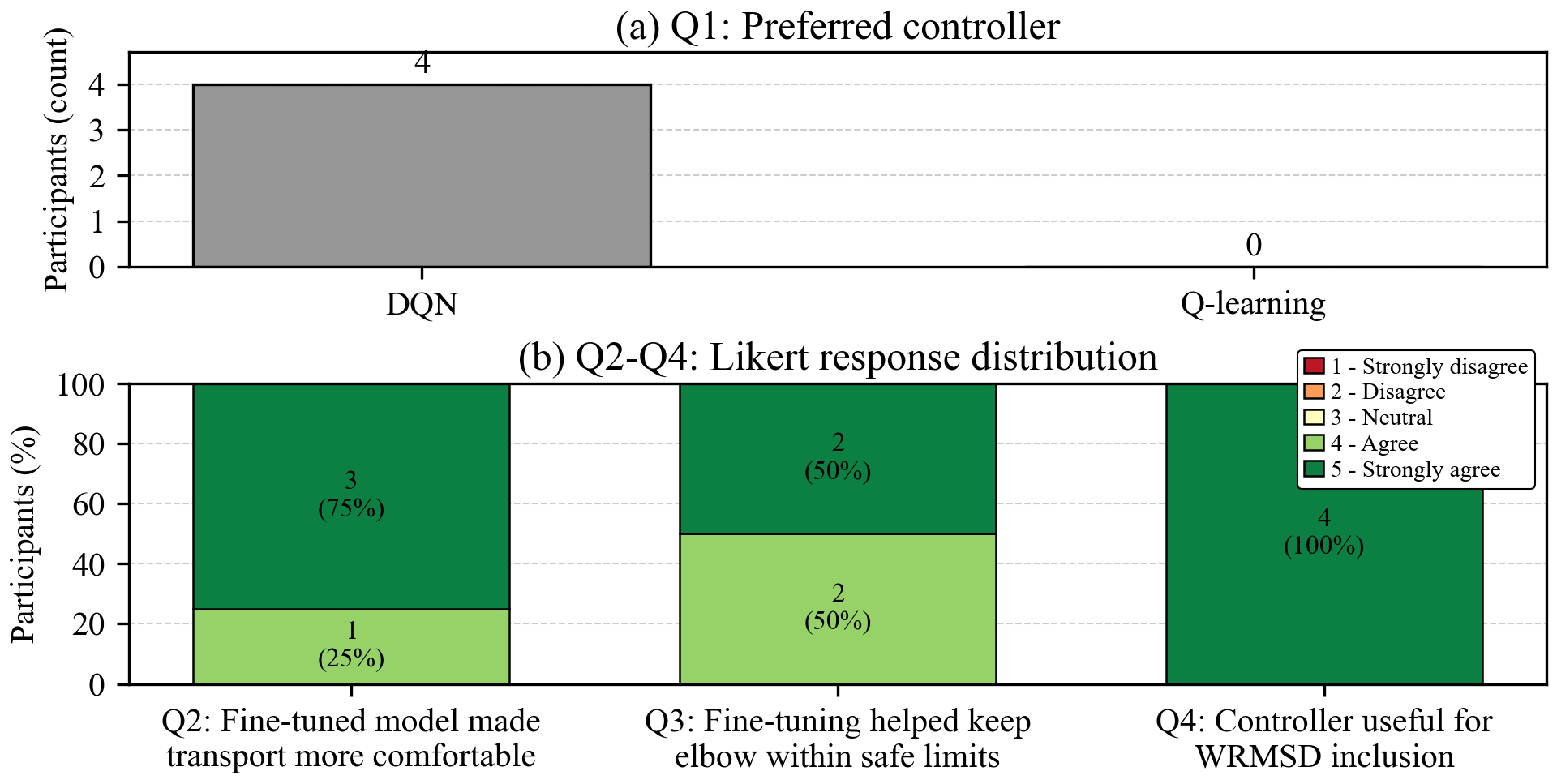}
    \caption{Usability questionnaire results ($n = 4$). Q1 shows participant preference between DQN and Q-learning (QL) in terms of perceived comfort.  Q2--Q4 report responses to usability statements in relation to DQN-based controller, on a 5-point Likert scale (1: strongly disagree, 5: strongly agree).  Percentages indicate the proportion of participants per response.}
    \label{fig:usability}
\end{figure}

The successful completion of the testing protocol by all participants underscores the controller’s strong adaptability across varying human anthropometries. This adaptability was further supported by the usability questionnaire, which reflected consistently positive user feedback.
Fig. \ref{fig:usability} presents the distribution of participants' responses to the usability questionnaire. Participants reported greater comfort during transport with the DQN method, noting that the fine-tuning phase effectively bridged the simulation-to-real gap. They also expressed confidence in the system’s applicability to industrial settings.

The proposed scheme shows real-world scalability by using simulation pre-training to establish an initial policy before deployment, reducing the amount of real-world interaction data required. As the state space grows (e.g., extending from a 2D Cartesian (x, z) space to 3D (x, y, z)), tabular Q-learning becomes impractical due to the exponential increase in state-action combinations. This leads to poor sample efficiency, as many interactions are required to explore the expanded space. In contrast, DQN enables generalization across states through function approximation, making it more suitable for high-dimensional problems.
Additionally, the human kinematic model supports seamless extension from 2D to 3D state spaces. By incorporating the y-axis into the Cartesian (x, z) framework and augmenting the model with additional degrees of freedom, the system enables this transition with minimal adaptation to the inverse kinematics formulation.

The results reinforce the effectiveness of fine-tuning in eliminating the simulation-to-real gap. Given the predefined requirements, the DQN model emerges as the preferable choice. Despite a longer pre-training phase due to its added complexity, DQN achieved superior transport efficiency compared to the Q-Learning method.
A demonstration video of the proposed system is available as supplementary material.

\section{Conclusion and Future Work} \label{sec: Conclusion and Future Work}
This study extends previous work by scaling to a larger and diverse participant group, with varying anthropometrics, and by introducing Deep Q-Networks as a more efficient alternative to Q-Learning, enabling a direct performance comparison between the two algorithms.
The learned policies successfully met the predefined solution requirements across both training phases. While a simulation-to-real gap was consistently observed, marked by reward drops and emerging pain risk, fine-tuning in the real environment successfully resolved these issues, allowing the agent to adapt to individual users.

DQN also outperformed Q-Learning by completing the transport task in fewer steps and, consequently, less time. Its flexible action and state space design that enabled variable step sizes, led to more efficient policies and faster convergence, making it a strong candidate for real-world deployment.
From a methodological perspective, employing both Q Learning and DQN revealed a clear trade off between training cost and model performance. Q Learning has short pre training times and require minimal computational resources, but its discrete state-action representation constrains trajectory efficiency and scalability. In contrast, DQN benefits from a continuous state representation and variable step sizes, which enabled shorter trajectories and reduced task time, at the cost of substantially longer initial pre training and the need for extensive hyperparameter tuning. 

While tabular Q-Learning offers conditional convergence guarantees, these are difficult to preserve in real-world physical human-robot interaction, where human motor variability, sensor noise, and user adaptation introduce non-stationarity that violates the underlying theoretical assumptions. DQN, by contrast, offers no formal convergence guarantees by design, as neural network function approximation inherently trades theoretical stability for generalization capacity. In both cases, learned Q-values should be interpreted as empirical estimates valid for the observed interaction distribution, rather than globally optimal solutions. This reflects an inherent trade-off in human-in-the-loop reinforcement learning: incorporating real human interaction is essential for obtaining transferable and meaningful policies, but comes at the cost of formal optimality.

Moreover, the performance of both controllers is inherently limited by the fidelity of the human kinematic model and the sensing system, as model simplifications and motion capture inaccuracies contribute to the simulation-to-real gap. Closing this gap requires real-world fine-tuning that consumes time, experimental resources, and participant effort. These costs should be weighed against the ergonomic and inclusivity gains when deploying such controllers in industrial settings.

While this exploratory study with four healthy participants successfully demonstrated adaptability across diverse anthropometries, the limited sample size precludes rigorous statistical analysis. Future work will extend testing to larger participant groups including WRMSD patients and replace the current machine-state based pain assessment with a digital pain biomarker, improving generalizability across musculoskeletal conditions. Additionally, incorporating a more realistic human kinematic model into the simulation may help further reduce the simulation-to-real gap. These efforts aim to create a safer, more adaptable controller for real-world deployment.

\bibliographystyle{IEEEtran}
\bibliography{biblio}

\end{document}